\documentclass{article}

\usepackage{PRIMEarxiv}
\usepackage[utf8]{inputenc}
\usepackage[T1]{fontenc}
\usepackage{amsmath,amsfonts,amssymb}
\usepackage{algorithmic}
\usepackage{algorithm}
\usepackage{array}
\usepackage{booktabs}
\usepackage[caption=false,font=normalsize,labelfont=sf,textfont=sf]{subfig}
\usepackage{textcomp}
\usepackage{stfloats}
\usepackage[hyphens]{url}
\usepackage{graphicx}
\usepackage{cite}
\usepackage{microtype}
\usepackage{hyperref}
\hypersetup{
  hidelinks,
  pdftitle={Disentangle and Drop: Robust Universal Removal of Image Watermarks via Reconstructive Grayscale Residual Decomposition},
  pdfauthor={Qi Li, Jidong Yang, Feng-Lei Fan, Yuantian Miao, Xiao Chen, Huaike Yu, Chunpeng Wang, Suo Gao, Herbert Ho-Ching Iu, and Bin Ma},
  pdfkeywords={invisible watermarking, watermark attacks, watermark removals, diffusion model, digital image watermarking}
}
\graphicspath{{./figures/out/}}
\newcommand{\method}{DnD}
\newcommand{\enc}{E}
\newcommand{\dec}{D}
\newcommand{\refine}{R}
\newcommand{\ag}{A_g}
\newcommand{\ax}{A_x}
\newcommand{\ber}{\operatorname{BER}}
\newcommand{\psnr}{\operatorname{PSNR}}
\newcommand{\ssim}{\operatorname{SSIM}}
\begin{document}

\title{Disentangle and Drop: Robust Universal Removal of Image Watermarks via Reconstructive Grayscale Residual Decomposition}

\author{%
\begin{minipage}{0.96\textwidth}
\centering
Qi Li$^{1}$, Jidong Yang$^{1,*}$, Feng-Lei Fan$^{2}$, Yuantian Miao$^{3}$, Xiao Chen$^{3}$, Huaike Yu$^{1}$,\\
Chunpeng Wang$^{1}$, Suo Gao$^{4}$, Herbert Ho-Ching Iu$^{5}$, and Bin Ma$^{1}$\\[0.8em]
{\normalfont\small
$^{1}$Key Laboratory of Computing Power Network and Information Security, Ministry of Education; Shandong Computer Science Center; Shandong Provincial Key Laboratory of Industrial Network and Information System Security; and Shandong Fundamental Research Center for Computer Science, Qilu University of Technology (Shandong Academy of Sciences), Jinan 250353, China\\
$^{2}$Department of Data Science, City University of Hong Kong, Hong Kong, China\\
$^{3}$School of Computer and Information Sciences, College of Engineering, Science and Environment, The University of Newcastle, Callaghan, NSW 2308, Australia\\
$^{4}$School of Information Science and Engineering, Dalian Polytechnic University, Dalian 116034, China\\
$^{5}$School of Electrical, Electronic and Computer Engineering, The University of Western Australia, Perth, WA 6009, Australia\\[0.4em]
\texttt{qluliqi@163.com; jidong\_yang\_paper@163.com; fenglfan@cityu.edu.hk; sky.miao@newcastle.edu.au}\\
\texttt{xiao.chen@newcastle.edu.au; huaikeyu@gmail.com; mpeng1122@163.com; gaosuodlpu@163.com}\\
\texttt{herbert.iu@uwa.edu.au; sddxmb@126.com}\\[0.3em]
$^{*}$Corresponding author: Jidong Yang.}
\end{minipage}%
}

\maketitle

\begin{abstract}
Invisible image watermarks are commonly evaluated against benign postprocessing operations such as compression, resizing, blur, and color changes. These tests leave out a different threat: a learned remover that preserves semantic image content while discarding residual evidence that carries the payload. We propose Disentangle and Drop (\method), an attack that is agnostic to the watermark method and treats watermark removal as a representation routing problem. \method{} decomposes a watermarked image into a semantic grayscale carrier and an auxiliary residual branch, and then suppresses the residual branch to reduce watermark evidence. The model is trained with latent spectral perturbations and low-strength diffusion exposure so that the drop operation remains stable under adaptive reconstruction. Experiments on seven representative watermark families show that one shared operating setting gives competitive removal with high visual fidelity. Operating scans and ablations separate usable attacks from image-damaging settings: stronger noise or diffusion can raise removal scores by damaging the image, while the practical regime comes from dropping the residual latent. These results argue for evaluating watermark robustness against learned removal at the representation level, not only against conventional image edits.
\end{abstract}


\section{Introduction}

Invisible image watermarking provides a technical mechanism for content provenance, copyright tracing, and accountability in generated media. Classical spread-spectrum and transform-domain schemes embed weak signals into image components with high perceptual tolerance \cite{podilchuk2001digital,hartung1999multimedia,cox1997secure}. Modern learned systems extend this principle with neural encoders, self-supervised feature spaces, and mechanisms designed for generative models \cite{zhu2018hidden,fernandez2022sslwatermarking,zhang2019rivagan,tancik2020stegastamp,fernandez2023stable,wen2023treering,gowal2025synthid,sander2025watermark}. Across these families, a common robustness claim is that the hidden payload should remain detectable after compression, resizing, color changes, mild blur, or other benign edits.

Robustness to benign edits is different from robustness to a learned remover. A watermark that survives JPEG compression, mild blur, resizing, or color shifts may still be vulnerable if a model can separate the semantic image carrier from residual evidence that carries the payload. Recent benchmarks and attack studies show that modern watermarks can fail under adaptive reconstruction, controllable regeneration, or diffusion removal tailored to a watermark even when they survive standard image edits \cite{an2024waves,zhao2024provably,liu2025ctrlregen,hu2024stableunstable,alam2025sadre}. However,these attack assume the adversories obtained some prior knowledge which motivates a more direct question: if the attacker does not know the watermark key, payload, embedder, or decoder, can a learned remover at the representation level attack a broad set of watermark families while preserving image contentS?

We study this setting by formulating watermark removal as a latent representation routing problem focusing on latent representation operation. The key observation is that many invisible watermarks occupy a small but detectable residual channel relative to image semantics. Instead of learning a direct purifier in image space, we propose \method{} operates on the latent representation produced by an encoder--decoder pair. It routes the visible image structure needed for reconstruction into a single-channel grayscale-like structural latent $g$, and routes payload-bearing and color residual evidence into an auxiliary latent $u$. Given a watermarked image $x_w$, \method{} computes $(g_w,u_w)=\enc(x_w)$, applies latent spectral perturbation $\hat g_w=\ag(g_w;K)$ to stress the carrier, and removes the watermark by decoding with the residual branch suppressed:
\begin{equation}
    x_0 = \dec(\hat g_w, 0).
\end{equation}
This ``drop'' operation is intentionally simple, making the attack inspectable, controllable, and easy to stress test.

The simple drop decoder, however, can be brittle. If a defender or evaluator applies spectral filtering to the latent representation or regenerates the image with a diffusion model, the payload can be reappeared, image quality can be degraded, or both. Therefore \method{} includes two adversarial operators, a latent operator and an image operator, during training. The latent operator $\ag$ attenuates low, mid, and high frequency bands of $g$ and injects magnitude and phase noise. The image operator $\ax$ performs img2img regeneration with diffusion and straight-through gradients, and a residual refiner $\refine$ repairs the resulting image using local restoration and attention context blocks. Herein, diffusion is used to stress and refine the drop at the representation level so that the final operating point does not rely on regeneration that visibly changes the image \cite{zhao2024provably,liu2025ctrlregen,zhang2024zodiac,alam2025sadre}.

The paper makes four contributions:
\begin{enumerate}
    \item It formulates a universal image watermark removal attack as a reconstructive grayscale residual representation routing problem in which visible semantics are retained in $g$ , and discardable residual evidence is suppressed by decoding from the routed structure latent with the auxiliary branch reduced or removed.
    \item It introduces a staged adversarial curriculum that combines latent spectral perturbations, diffusion regeneration, and a residual refiner trained through a straight-through path.
    \item It provides a formal and empirical analysis suite for removability, including an attack definition with a fidelity constraint, an information gate condition, counterfactual latent swaps, pixel energy decomposition, decoder sensitivity probes, spectral localization, and a Fano-type payload information certificate.
    \item It evaluates one shared $K/U/S$ operating point across seven watermarking methods: SSL, DwtDct, DwtDctSvd, RivaGAN, StegaStamp, Stable Signature, and Tree-Ring.
\end{enumerate}

\section{Related Work}

\subsection{Watermarking}
Classical digital watermarking embeds signals in frequency or spread-spectrum coordinates and often balances robustness with perceptual distortion through transform-domain design \cite{podilchuk2001digital,hartung1999multimedia,cox1997secure}. Practical libraries expose DWT/DCT and DWT/DCT/SVD variants, which remain useful baselines because their payloads are spatially distributed but frequency localized \cite{barni2001wavelet,barni1998dct,kumari2023dwtsvd}. Learned watermarking methods use neural encoders and decoders to hide payloads in image residuals. HiDDeN demonstrates end-to-end data hiding with neural networks \cite{zhu2018hidden}; StegaStamp targets robust decoding after display and photography \cite{tancik2020stegastamp}; RivaGAN adds attention and adversarial critics for robust invisible video watermarking \cite{zhang2019rivagan}. SSL Watermarking embeds messages in self-supervised latent spaces \cite{fernandez2022sslwatermarking}. Generative watermarking systems such as Stable Signature, Tree-Ring, SynthID Image, and localized message schemes integrate watermark signals into generated media pipelines \cite{fernandez2023stable,wen2023treering,gowal2025synthid,sander2025watermark}. Recent diffusion-native and provenance-oriented watermarking systems further move watermark evidence from post-hoc pixel residuals to initial noise, latent distributions, semantic attribution, or localized message regions, including Gaussian Shading, ProMark, RingID, two-stage noise-based watermarking, and arbitrary-resolution watermarking systems \cite{yang2024gaussianshading,asnani2024promark,ci2024ringid,arabi2025hidden,bui2025trustmark}. This variety matters for removal evaluation: robustness to a standard suite of image edits does not imply that the payload is inseparable from image semantics.

\subsection{Removal and Regeneration Attacks}
A watermark can fail under ordinary image processing, but a security evaluation should also consider adaptive removal. Recent robustness benchmarks emphasize that a watermark should be tested against attack families beyond benign postprocessing \cite{an2024waves}. Regeneration attacks weaken hidden payloads and reconstruct the visible image using generative priors \cite{zhao2024provably,liu2025ctrlregen}. Removal specific to Stable Signature and attack-resilient diffusion watermarking further show that the interaction between generative priors and hidden signals is central to security \cite{hu2024stableunstable,zhang2024zodiac}. SADRE introduces saliency-aware diffusion reconstruction, combining adaptive noise injection with reverse diffusion to balance watermark disruption and perceptual fidelity \cite{alam2025sadre}. Beyond regeneration attacks, recent no-box and challenge-driven studies show that watermark detectors can also be weakened through transferable perturbations or adaptive restoration pipelines without full access to the target detector, which further motivates evaluating learned removers under weaker attacker knowledge assumptions \cite{hu2025transferattack}. Together, these studies make diffusion regeneration and no-box transfer attacks demanding adaptive tests for watermark removers. \method{} differs by learning a reusable representation split for each image before regeneration: it preserves the semantic carrier in $g$ and suppresses a discardable residual branch $u$, rather than relying on diffusion alone as the remover.

\subsection{Representation Reconstruction}
The Stage 3 refiner in \method{} follows a practical lesson from image restoration: local convolutions are efficient, but perceptual and frequency-aware reconstruction losses help repair structured degradation. VGG features are widely used as perceptual anchors for image transformation, and focal frequency loss motivates explicit matching of spectral structure during reconstruction \cite{simonyan2015vgg,johnson2016perceptual,jiang2021focal}. The representation split also connects to representation learning, information bottleneck, and disentanglement: a useful latent should preserve visible content relevant to the task while suppressing nuisance evidence that can be routed elsewhere \cite{bengio2013representation,tishby2000information,alemi2017deep,locatello2019challenging}. In \method{}, this principle is used as a prior for watermark removal, not as a decoder specific to a watermark.

\subsection{Adaptive Defenses}
Defenses against learned removers can make the watermark less separable from the visible image or can reconstruct the image after an attack. Two directions are especially relevant to the threat model studied here. The first strengthens robustness in spectral coordinates, following the intuition behind spread spectrum, transform domain, and frequency-aware learned watermarks that payload evidence should survive common filtering. The second uses regeneration or purification, often with diffusion models, to reconstitute the visible image and suppress suspicious residual perturbations. Recent defenses also move toward semantic and editing-aware training, certified robustness, generative-prior robustness, and undetectable latent watermarking, suggesting that future watermark designs should reduce residual separability rather than only increase robustness to conventional pixel-level distortions \cite{hu2024robustwide,jiang2024certifiably,lu2025vine,gunn2025undetectable}. We select these two defenses because they test the two assumptions behind \method{}: the semantic carrier $g$ should remain stable under latent spectral perturbation, and the residual drop should remain effective after adaptive image reconstruction. They also cover practical countermeasures available to a black-box defender without assuming knowledge of the remover parameters.

\section{Threat Model and Objective}

The attacker receives only a watermarked RGB image $x_w\in[-1,1]^{3\times H\times W}$, where $H$ and $W$ denote image height and width. The attacker does not know the watermark method $m$, decoder $\operatorname{Dec}_m$, embedding key, or payload bits. The goal is to produce an output $\mathcal{A}(x_w)$ that preserves visible content while making the payload unreadable. For multi-bit watermarks, success is measured by an increased bit error rate (BER); for zero-bit or detector-based watermarks such as Tree-Ring, success is measured by reduced detection at a fixed false positive rate.

For multi-bit payloads, the ideal attack objective is written as
\begin{equation}
\begin{aligned}
\max_{\theta}\quad
&\mathbb{E}_{x_w}\!\left[
    \min\{\ber(\operatorname{Dec}_m(\mathcal{A}_{\theta}(x_w))),0.5\}
\right]\\
\text{s.t.}\quad
&\mathbb{E}_{x_w}\!\left[d(\mathcal{A}_{\theta}(x_w),x_c)\right]\le \tau .
\end{aligned}
\end{equation}
Here, $x_c$ is the paired clean image when available, $d$ is a perceptual distortion measure, and $\tau$ sets the fidelity constraint. The cap at 0.5 reflects the random guessing ceiling for uniformly distributed bits; raw BER is still reported in the experiments. For detector-based watermarks, the BER term is replaced by the corresponding detector objective, such as reducing acceptance at the fixed false positive rate operating point. In this sense, a successful attack should leave little payload information recoverable by the evaluated decoder while preserving the visible image content. We formalize this objective below as a security criterion with a fidelity constraint rather than as an unconstrained image degradation task.

We consider two adaptive defenses. A defender may make payloads robust to spectral filtering or hide them in frequency bands that survive common operations. A defender may also use diffusion regeneration to purify or reconstitute the watermarked image. \method{} is trained against both cases.

Table~\ref{tab:notation} summarizes the notation used in the method and analysis sections.

\begin{table*}[t]
\caption{Summary of notation. Subscripts $w$ and $c$ denote watermarked and clean carriers when both are available.}
\label{tab:notation}
\centering
\footnotesize
\setlength{\tabcolsep}{3pt}
\begin{tabular}{@{}>{\raggedright\arraybackslash}p{0.15\textwidth}>{\raggedright\arraybackslash}p{0.32\textwidth}>{\raggedright\arraybackslash}p{0.15\textwidth}>{\raggedright\arraybackslash}p{0.32\textwidth}@{}}
\toprule
Symbol & Meaning & Symbol & Meaning \\
\midrule
$x_w$ & watermarked RGB input image & $H,W$ & image height and width \\
$x_c$ & clean carrier or paired clean image when available & $m$ & watermarking method under evaluation \\
$\operatorname{Dec}_m$ & method-specific watermark decoder or detector & $\mathcal{A},\mathcal{A}_{\theta}$ & attack mapping and its parameterized form \\
$d,\tau$ & distortion measure and fidelity constraint threshold & $\theta$ & parameters of the attack mapping \\
$\enc,\dec$ & \method{} encoder and decoder & $g,u$ & structure latent and auxiliary residual latent \\
$(g_w,u_w)$ & latent pair encoded from $x_w$ & $(g_c,u_c)$ & latent pair encoded from $x_c$ when available \\
$\hat g_w$ & structure latent after latent spectral perturbation & $g_A$ & structure latent encoded from $x_A$ \\
$f_g,f_u$ & structure and auxiliary encoder heads & $\ag(g;K)$ & latent spectral perturbation applied to $g$ with strength $K$ \\
$\ax(x;S)$ & diffusion stress operator applied to decoded image $x$ with strength $S$ & $\refine(x_A,g_A)$ & residual refiner conditioned on $x_A$ and $g_A$ \\
$K$, $U$, $S$ & latent perturbation strength, auxiliary keep ratio, and diffusion strength & $x_0$ & strict drop-decoded image $\dec(\hat g_w,0)$ \\
$x_0^{(U)}$ & controlled K/U decoded image $\dec(\hat g_w,Uu_w)$, with $x_0=x_0^{(0)}$ & $x_A$ & diffusion-stressed image, $\ax(x_0;S)$ for strict inference or $\ax(x_0^{(U)};S)$ for scans \\
$x_R$, $x_{\mathrm{out}}$ & strict refined image and final attacked output; superscript $(U)$ denotes the scan endpoint & $\tilde{x}_A$ & straight-through diffusion image used in Stage 3 training \\
$\Omega_{\ell}$, $\Omega_m$, $\Omega_h$ & low, mid, and high spectral bands & $r$ & normalized radial frequency coordinate \\
$\kappa_{\ell,m,h}^{(s)}$ & stage-dependent keep ratios for spectral bands at stage $s$ & $N,G$ & diffusion step count and guidance scale \\
$\epsilon_g,\eta_g,\delta,\rho$ & structure stability and residual gate quantities in Proposition 1 & $r_\phi,h,\alpha$ & refiner residual network, high-frequency residual extractor, and residual scale \\
$ww$, $wc$, $cw$, $w0$ & latent swap counterfactuals & $E_W$, $E_g$, $E_u$ & pixel energy terms for watermark, structure, and residual branches \\
$b,\hat b,p,h_2$ & payload bit, decoded bit, BER, and binary entropy in the Fano certificate & $\ber$, $\psnr$, $\ssim$ & removal and fidelity metrics \\
\bottomrule
\end{tabular}
\end{table*}

\section{Method}

\subsection{Overview}

\method{} is built around a representation routing view of watermark removal. Instead of treating every pixel as equally suspect, it asks whether the visible image carrier and the watermark residual evidence can be assigned different roles in the latent space. The structure latent $g$ is expected to retain the semantic carrier needed for reconstruction, while the auxiliary latent $u$ is encouraged to absorb residual evidence that can be removed without destroying image content. Removal is therefore expressed as an intervention on the representation: keep the carrier, suppress the residual branch, and decode the image from the routed representation.

\begin{figure*}[!t]
\centering
\includegraphics[width=0.93\textwidth]{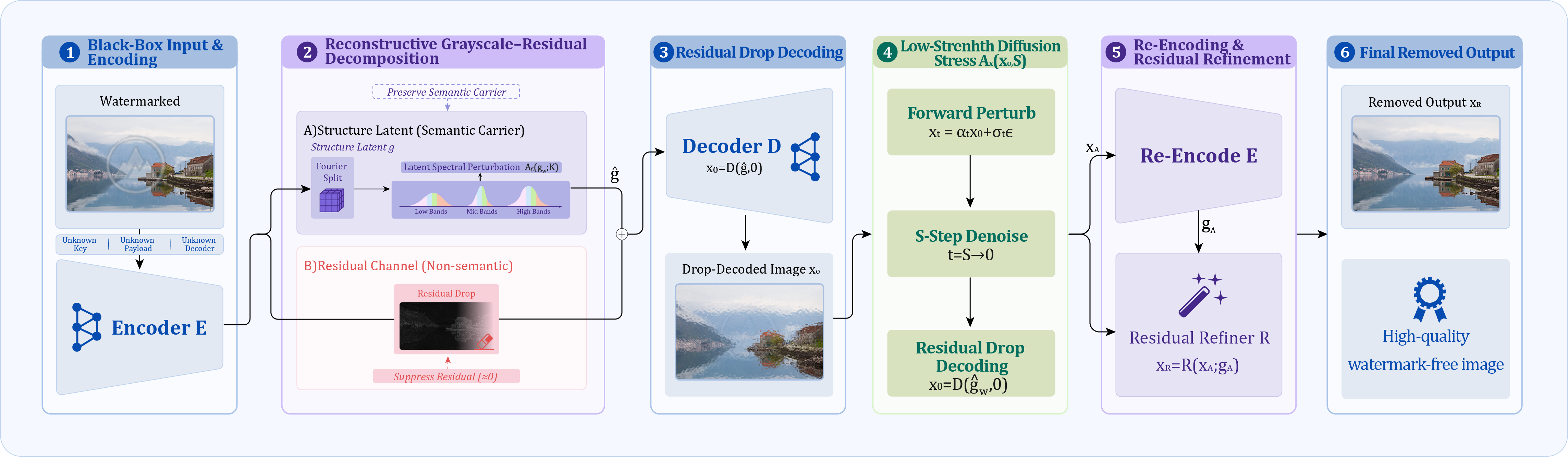}
\caption{Overview of the \method{} inference pipeline. A watermarked image is encoded into a structure latent $g$ and an auxiliary residual latent $u$. The structure branch is preserved under the latent spectral perturbation $\ag(g;K)$, while the auxiliary residual branch is suppressed before decoding. The dropped output $x_0$ can then pass through low-strength diffusion stress $\ax(x_0;S)$ and residual refinement $\refine$ to produce the final output $x_R$.}
\label{fig:pipeline}
\end{figure*}

Figure~\ref{fig:pipeline} summarizes this inference logic. Given a watermarked image, \method{} first encodes it into $(g_w,u_w)$. A latent spectral perturbation $\ag(g_w;K)$ stresses the structure branch so that the carrier remains stable under spectral variation. The residual branch is then dropped by setting $u\rightarrow 0$, yielding a drop-decoded image. Low-strength diffusion stress $\ax(x_0;S)$ and the residual refiner $\refine$ are applied after this drop as stress and repair components, rather than as the main source of watermark removal. This design keeps the attack compatible with the black-box threat model while making carrier preservation, residual suppression, and repair after dropping explicit.

\method{} uses five operators:
\begin{equation}
    \enc(x)=(g,u),\quad \ag(g;K),\quad \dec(g,u),
\end{equation}
\begin{equation}
    \ax(x;S),\quad \refine(x_A,g_A).
\end{equation}
The strict inference path is
\begin{equation}
\begin{aligned}
(g_w,u_w) &= \enc(x_w),\\
\hat g_w &= \ag(g_w;K),\\
x_0 &= \dec(\hat g_w,0),\\
x_A &= \ax(x_0;S),\\
(g_A,\_) &= \enc(x_A),\\
x_R &= \refine(x_A,g_A).
\end{aligned}
\end{equation}
The final attacked output is
\begin{equation}
x_{\mathrm{out}} =
\begin{cases}
x_R, & S>0,\\
x_0, & S=0,
\end{cases}
\qquad
\mathcal{A}_{\theta}(x_w;K,0,S)=x_{\mathrm{out}}.
\end{equation}
For operating scans, we separately define a controlled K/U decoded image
\begin{equation}
    x_0^{(U)}=\dec(\hat g_w,Uu_w),\qquad x_0=x_0^{(0)}.
\end{equation}
The evaluated scan endpoint then follows the same stress and repair logic:
\begin{equation}
\begin{aligned}
x_A^{(U)} &= \ax(x_0^{(U)};S),\\
(g_A^{(U)},\_) &= \enc(x_A^{(U)}),\\
x_R^{(U)} &= \refine(x_A^{(U)},g_A^{(U)}).
\end{aligned}
\end{equation}
The evaluated attacked image for a scan point is
\begin{equation}
x_{\mathrm{out}}^{(U)} =
\begin{cases}
x_R^{(U)}, & S>0,\\
x_0^{(U)}, & S=0,
\end{cases}
\qquad
\mathcal{A}_{\theta}(x_w;K,U,S)=x_{\mathrm{out}}^{(U)}.
\end{equation}
Thus the strict algorithm is the special case $U=0$, while the reported $K/U/S$ operating points use $U$ as a controlled residual keep ratio. $K$ controls the strength of latent spectral perturbation, $U$ controls how much of the auxiliary residual is reintroduced before diffusion, and $S$ controls the strength of diffusion stress. During inference, $S=0$ disables diffusion, and the refiner is skipped unless it is explicitly requested. The remainder of this section first defines the residual decomposition and its removability criterion, and then describes the spectral stress, diffusion stress, residual refinement, and training curriculum.

\subsection{Reconstructive Grayscale Residual Decomposition}

The encoder firstly takes the watermarked image's RGB channels together with a high-frequency grayscale residual, forming a four-channel input. A U-Net-like residual backbone maps this input to shared features. When enabled, a Fourier disentanglement block applies rFFT, grouped channel mixing over real and imaginary components, soft-shrinkage, and inverse rFFT before the final heads to extract a structure latent. Afterwards, the heads produce a structure latent with one channel
\begin{equation}
    g=\tanh(f_g(x))
\end{equation}
and a 16-channel auxiliary latent
\begin{equation}
    u=\tanh(f_u(x)).
\end{equation}
Here, $f_g$ and $f_u$ denote the structure and auxiliary encoder heads, respectively.
The decoder concatenates $g$, $u$, and a high-frequency residual extracted from $g$, to reconstruct the RGB values with residual convolutional blocks. This design separates a branch that preserves structure from an auxiliary residual branch.

The design biases $g$ toward a natural grayscale carrier. If the full pair $(g_w,u_w)$ reconstructs $x_w$ while $\dec(g_w,0)$ moves toward $x_c$, then the payload has been routed to the discardable branch. This decomposition and representation routing behavior is the core mechanism leveraged by DnD, allowing watermark-bearing features to be concentrated in a suppressible branch while semantic image content remains largely intact.

\subsection{Formal Analysis of Residual Channel Removability}

Rather than relying on indiscriminate image degradation, DnD treats watermark removal as a representation-routing problem. Under a prescribed image-quality constraint, the attack suppresses the representation branch that predominantly carries watermark-related residual information while preserving semantic image content. The effectiveness of the attack is then evaluated by determining whether any watermark evidence remains recoverable. This perspective naturally motivates the following theoretical analysis, which formalizes watermark removability through latent representation separability and information localization.

Mutual information is used here as an explanatory quantity to characterize the dependence between the residual representation and the embedded watermark. When direct estimation is required, practical neural and variational estimators can be employed \cite{belghazi2018mine,poole2019variational}.

\noindent Definition 1 (Watermark removal under a fidelity constraint).
Given a watermarked image $x_w$, an attack $\mathcal{A}$ is successful under the fidelity constraint $\tau$ if it reduces the watermark verification signal while satisfying
\begin{equation}
    d(\mathcal{A}(x_w),x_c)\le \tau .
\end{equation}
For bit payloads, reducing verification means driving the BER measured by the watermark decoder $\operatorname{Dec}_m$ toward random guessing; for detector-based watermarks, it means reducing the acceptance rate of the corresponding detector at the evaluation threshold. The definition makes fidelity part of the security criterion, so an output that visibly damages the image is not counted as a useful removal unless it satisfies the constraint.

\noindent Definition 2 (Residual channel separability).
Let $\enc(x_w)=(g_w,u_w)$ and let $\dec$ be the corresponding decoder. A watermarked image is separable through a residual channel under $(\enc,\dec)$ when $g_w$ preserves the semantic carrier and $u_w$ contains the residual evidence needed to reconstruct the watermark-bearing observation:
\begin{equation}
    \dec(g_w,u_w)\approx x_w,\qquad
    \dec(g_w,0)\approx x_c .
\end{equation}
This definition gives $g$ and $u$ an operational security meaning: $g$ should support visible reconstruction, while $u$ should carry evidence that can be suppressed without violating the fidelity constraint.
These separability statements are written for the unperturbed latent pair. The inference path later replaces $g_w$ with $\hat g_w=\ag(g_w;K)$ so that the same drop operation is tested under latent spectral stress.

\noindent Proposition 1 (Information gate removability condition).
Let
\begin{equation}
\begin{aligned}
\epsilon_g&=\operatorname{RMS}(g_w-g_c),\\
\delta&=\operatorname{RMS}(\dec(g_w,u_w)-x_w),\\
\rho&=\operatorname{RMS}(\dec(g_w,0)-x_w).
\end{aligned}
\end{equation}
Assume that $\dec(g_w,u_w)$ reconstructs the watermarked observation, that $\dec(g_w,0)$ satisfies the fidelity constraint relative to the clean carrier, that the structure branch remains close to the clean carrier in latent space for a small tolerance $\eta_g$, and that dropping the auxiliary branch separates the output from the watermarked observation:
\begin{equation}
    \epsilon_g \le \eta_g,\qquad \rho-\delta>0 .
\end{equation}
Then information absent from $g_w$ but present in $u_w$ is required to reconstruct the watermark-bearing observation. Suppressing $u_w$ is a valid removal operation through the residual channel under the fidelity constraint.

\noindent Proof sketch.
The condition $\epsilon_g \le \eta_g$ checks that the structure latent stays close to the clean carrier, so the structure branch is not the main source of watermark-specific residual variation. The full latent pair reconstructs $x_w$ with error $\delta$, while the dropped pair is farther from $x_w$ by $\rho$. If the dropped output still preserves the clean carrier, the missing contribution is residual evidence needed to reproduce the watermarked observation. The joint gate $(\epsilon_g,\rho-\delta)$ is thus a falsifiable condition for removability rather than merely a diagnostic score.
This certificate is written for the unperturbed latent pair. The deployed attack replaces $g_w$ with $\hat g_w=\ag(g_w;K)$, and the stability of this perturbed drop path is checked by the spectral localization probe and the operating scans.

\noindent Lemma 1 (Counterfactual localization of payload evidence).
Consider the latent swaps
\begin{equation}
\begin{aligned}
ww&=\dec(g_w,u_w),&
wc&=\dec(g_w,u_c),\\
cw&=\dec(g_c,u_w),&
w0&=\dec(g_w,0).
\end{aligned}
\end{equation}
If replacing $u_c$ with $u_w$ increases payload recoverability, while replacing $u_w$ with zero suppresses recoverability, then the evaluated payload evidence is more strongly localized in the auxiliary branch than in the structure branch.

\noindent Proof sketch.
The counterfactuals hold the decoder fixed and exchange only one branch at a time. A payload readout that follows $u_w$ across swaps and disappears under $u=0$ is evidence that the auxiliary latent, rather than the grayscale structure latent alone, carries the evaluated watermark signal.

\noindent Proposition 2 (Payload information certificate at the decoder level).
For a uniformly distributed binary payload bit $b$ and a decoder output $\hat b$ with bit error rate $p$, a Fano-type bitwise certificate is
\begin{equation}
    I(b;\hat b) \ge 1-h_2(\min(p,0.5)),
\end{equation}
where $h_2$ is binary entropy. As $p$ approaches $0.5$, this lower bound certificate at the decoder level vanishes. This result should be read conservatively: it shows that the evaluated decoder no longer certifies recoverable payload information, not that all possible information about the payload has been erased.

\noindent Corollary 1 (Removal under a fidelity constraint is governed by latent control).
Under a fixed fidelity constraint, useful removal is achieved when suppressing the auxiliary residual branch increases BER or detector removal while preserving PSNR and SSIM. Strong perturbation in image space or high diffusion strength may further increase removal metrics, but such gains correspond to low-fidelity settings when they violate the fidelity constraint. The practical attack is best interpreted as suppression of the residual channel, with diffusion serving as an adaptive stressor and low-strength refinement rather than as the core remover.

\subsection{Latent Spectral Perturbation for Carrier Stability}

The representation split is useful only if the structure latent remains a stable semantic carrier after watermark-related residual evidence has been routed away from it. A model trained only with the clean drop $\dec(g,0)$ can become brittle: small spectral changes in $g$ may either reintroduce residual evidence or damage the decoded image. We therefore apply a controlled latent spectral perturbation $\ag(g;K)$ to the structure branch. Its role is not to remove the watermark by corrupting the carrier, but to make the drop path rely on content that is stable across frequency bands.

The operator partitions the spectrum of $g$ into low, mid, and high radial bands, where $r$ denotes the normalized radial frequency:
\begin{equation}
    \Omega_{\ell}:r\le 1/3,\quad
    \Omega_m:1/3<r\le 2/3,\quad
    \Omega_h:r>2/3.
\end{equation}
For training stage $s$, each band has a target keep ratio $\kappa_{\ell,m,h}^{(s)}$, and the scalar strength $K$ interpolates between the identity mapping and these stage-dependent targets. The operator can additionally apply a ring notch, magnitude noise in the mid and high bands, and phase noise in the high band before reconstructing the latent with inverse rFFT. Because $\ag$ acts only on $g$, the decoder must preserve visible structure from a spectrally stable carrier while the auxiliary branch is suppressed. This makes $K$ an interpretable robustness control rather than a watermark-specific decoder or a pixel-space attack.

Algorithm~\ref{alg:dnd} summarizes the strict $U=0$ inference path after this latent perturbation is defined. The placement of the strength parameters is deliberate: $K$ affects the structure latent before residual dropping, while $S$ affects only the post-drop diffusion stress. Operating scans replace the strict decoded image $x_0$ by $x_0^{(U)}$ before applying the same diffusion and refiner path.

\begin{algorithm}[t]
\caption{\method{} Inference with Spectral and Diffusion Controls}
\label{alg:dnd}
Input: watermarked image $x_w$, strengths $K,S$\\
Output: attacked image $x_{\mathrm{out}}$
\begin{algorithmic}[1]
\STATE $(g_w,u_w)\leftarrow \enc(x_w)$
\STATE $\hat g_w \leftarrow \ag(g_w;K)$
\STATE $x_0 \leftarrow \dec(\hat g_w,0)$
\IF{$S>0$}
\STATE $x_A \leftarrow \ax(x_0;S)$
\STATE $(g_A,\_)\leftarrow \enc(x_A)$
\STATE $x_{\mathrm{out}}\leftarrow \refine(x_A,g_A)$
\ELSE
\STATE $x_{\mathrm{out}}\leftarrow x_0$
\ENDIF
\STATE return $x_{\mathrm{out}}$
\end{algorithmic}
\end{algorithm}

\subsection{Diffusion Regeneration and Straight Through Training}

The image stressor $\ax(x;S)$ uses Stable Diffusion img2img regeneration. It maps the selected decoded image to the diffusion domain, applies regeneration with strength $S$, step count $N$, and guidance $G$, and maps the result back to the image domain. In strict inference $x=x_0$; in operating scans $x=x_0^{(U)}$. Experiments use SD-2.1-base as the diffusion backbone. Let $x_A=\ax(x;S)$. To keep Stage 3 repair trainable, the diffusion output is inserted through a straight-through estimator:
\begin{equation}
    \tilde x_A = x + (x_A-x)_{\mathrm{detach}}.
\end{equation}
The forward value is the regenerated image, while gradients flow as if the operation were the identity mapping. The diffusion operator also supports residual scaling in low, mid, and high bands, as well as optional high-frequency detail preservation. These controls allow Stage 3 to start with a lighter regeneration setting and later evaluate stronger $S$ values.

\subsection{Residual Refiner}

The refiner $\refine$ is a residual restoration module:
\begin{equation}
    x_R = \operatorname{clip}(x_A+\alpha \tanh(r_\phi(x_A,h(x_A),g_A))).
\end{equation}
Here, $r_\phi$ is the refiner residual network with parameters $\phi$, $h(\cdot)$ extracts the high-frequency residual, and $\alpha$ scales the residual correction. The refiner receives the regenerated image, a high-frequency residual of that image, and the structure latent $g_A$ encoded again from the regenerated image. The architecture combines NAF-style local blocks with Restormer context blocks at a lower spatial resolution. The Stage 3 setting uses width 64, two Restormer blocks, four attention heads, and $\alpha=0.5$.

\subsection{Training Curriculum}

Training follows a three stage curriculum that first stabilizes reconstruction, then learns robust latent dropping, and finally trains repair under diffusion exposure.

\subsubsection{Stage 1: Reconstruction Warmup}
For 20 epochs, the model learns stable reconstruction of clean and watermarked images. The perturbation strength is zero, the drop loss is inactive, and the objective emphasizes reconstruction consistency and grayscale conformity.

\subsubsection{Stage 2: Latent Drop Robustness}
For 70 epochs, $\ag$ is enabled and the perturbation strength increases according to a cosine schedule. The drop objective $\|\dec(\ag(g_w;K),0)-x_c\|_1$ is warmed in from a smaller weight to the target weight. The watermarked reconstruction term decays to a nonzero floor, which preserves decoder capacity without allowing the auxiliary branch to dominate the structure branch.

\subsubsection{Stage 3: Diffusion Repair}
For 50 epochs, the Stage 2 encoder, decoder, and latent perturbation base are frozen by default. Diffusion regeneration is enabled with a light repair first setting ($S=0.10$, 6 steps, and guidance 1.0), and only the repair path is trained. The unified perturbation strength increases from 0 to 1 over 15 epochs, so the refiner is not initialized under the most aggressive regeneration setting.

\subsection{Loss Groups}

The training objective contains several scalar weights. They serve five roles. Fidelity anchors include clean reconstruction, watermarked reconstruction, and optional clean-drop anchoring. Grayscale conformity is encouraged through lightness, VGG contrast, and local total variation structure for $g$. Payload removal is trained through the direct $\ell_1$ drop loss, attacked latent alignment, and Stage 3 recovery loss. Perceptual and structural consistency are enforced with pooled VGG cosine similarity, multi-scale high-frequency loss, Sobel edge loss, and multi-band Fourier separation. Latent regularization uses clean auxiliary sparsity, a watermarked auxiliary floor, and quantization regularization for attacked $g$. Together, these roles balance reconstruction fidelity, payload removal, spectral robustness, and perceptual consistency.

\section{Why the Drop Works}

The drop attack is useful only if the learned representation separates visible structure from residuals that carry payload information. This requirement connects three observations from prior work. Robustness benchmarks show that benign postprocessing is not a sufficient security test \cite{an2024waves}. Regeneration attacks show that watermark disruption can be traded against perceptual fidelity through reconstruction \cite{zhao2024provably,liu2025ctrlregen,alam2025sadre}. Studies on representation learning and disentanglement suggest that useful factors can be routed into separate coordinates when the objective makes that separation operational \cite{bengio2013representation,locatello2019challenging}. The probes below operationalize the formal statements in the previous section: payload evidence should become measurable in a discardable channel, while the semantic carrier remains stable.

\subsection{Information Gate}
Let
\begin{equation}
\begin{aligned}
\epsilon_g&=\operatorname{RMS}(g_w-g_c),\\
\delta&=\operatorname{RMS}(\dec(g_w,u_w)-x_w),\\
\rho&=\operatorname{RMS}(\dec(g_w,0)-x_w).
\end{aligned}
\end{equation}
The gate condition is $\epsilon_g\le\eta_g$ together with $\rho-\delta>0$. The first term checks that the structure latent remains close to the clean carrier, while the second indicates that the full latent pair reconstructs the watermarked image more faithfully than the dropped pair. Together, they test whether information not present in $g$ is necessary to reproduce the watermark-bearing image.

\subsection{Counterfactual Payload Swaps}
Four counterfactuals are decoded:
\begin{equation}
\begin{aligned}
ww&=\dec(g_w,u_w),&
wc&=\dec(g_w,u_c),\\
cw&=\dec(g_c,u_w),&
w0&=\dec(g_w,0).
\end{aligned}
\end{equation}
If payload information is localized in $u$, then $cw$ should reintroduce payload evidence and $w0$ should suppress it. These counterfactual images are evaluated with the same decoders used for the main BER measurements.

\subsection{Pixel Energy Decomposition}
Finite-difference decoder swaps are used to estimate
\begin{equation}
\begin{aligned}
E_W&=\|\dec(g_w,u_w)-\dec(g_c,u_c)\|_2^2,\\
E_g&=\|\dec(g_w,u_c)-\dec(g_c,u_c)\|_2^2,\\
E_u&=\|\dec(g_c,u_w)-\dec(g_c,u_c)\|_2^2 .
\end{aligned}
\end{equation}
with cross term $E_W-E_g-E_u$. A successful split should show that a large share of pixel energy induced by the watermark can be attributed to $u$, while cross energy remains controlled.

\subsection{Decoder Sensitivity}
The Lipschitz probe estimates how much the decoder output changes under unit perturbations in $g$ and $u$. This separates a semantic branch that must be stable from an auxiliary branch that can be discarded.

\subsection{Spectral Localization}
The band epsilon probe measures low, mid, and high residuals for $g_w-g_c$ and $\ag(g_w;K)-g_c$ using the same radial bands as the spectral stressor. This probe tests whether the stressor aligns with the frequency bands in which watermark differences concentrate.

\subsection{Payload Information}
For a uniform binary payload and BER $p$ measured by the watermark decoder $\operatorname{Dec}_m$, a bitwise Fano-type lower bound is
\begin{equation}
    I(b;\hat b) \ge 1-h_2(\min(p,0.5)),
\end{equation}
where $h_2$ is binary entropy. Pushing BER toward 0.5 drives this certificate at the decoder level toward zero, which provides an information-theoretic interpretation of the loss of recoverability for the evaluated decoder without claiming that all possible payload information is absent.

Table~\ref{tab:residual-channel-analysis} summarizes how these probes close the loop between the formal residual channel claim and the measured behavior. The table is intended as a falsifiable analysis checklist: failure of any probe would weaken the interpretation that \method{} removes watermarks by suppressing a separable residual channel rather than by simply degrading the image.

\begin{table*}[t]
\caption{Residual channel separability and payload routing analysis. The probes map the formal removability argument to measurable predictions, separating targeted residual suppression from untargeted image degradation.}
\label{tab:residual-channel-analysis}
\centering
\small
\setlength{\tabcolsep}{4pt}
\resizebox{\linewidth}{!}{%
\begin{tabular}{llll}
\toprule
Probe & Formal role & Metric & Observed support \\
\midrule
B1 information gate & tests Proposition 1 & $(\epsilon_g,\rho-\delta)$ & small structure drift and positive gap with clean carrier fidelity preserved \\
B2 counterfactual swaps & tests Lemma 1 & BER or detector response of $cw$ and $w0$ & payload evidence follows $u_w$ and is suppressed by $u=0$ \\
B3 pixel energy & supports residual channel separability & $E_g,E_u,E_W-E_g-E_u$ & watermark induced energy is explained mainly by the auxiliary branch \\
B4 decoder sensitivity & checks branch geometry & unit-response sensitivity to $g$ and $u$ & $g$ remains stable while $u$ is discardable \\
B5 spectral localization & checks the latent stressor & low/mid/high residual bands & watermark residuals concentrate in bands targeted by $\ag$ \\
B6 information certificate & links BER to recoverability & $1-h_2(\min(p,0.5))$ & certificate decreases as BER approaches random guessing \\
\bottomrule
\end{tabular}%
}
\end{table*}

\section{Experimental Protocol}

\subsection{Dataset and Methods}
Experiments are conducted on paired clean and watermarked images resized to $512\times512$. The evaluation covers seven watermark families: SSL \cite{fernandez2022sslwatermarking}, DwtDct \cite{barni1998dct}, DwtDctSvd \cite{kumari2023dwtsvd}, RivaGAN \cite{zhang2019rivagan}, StegaStamp \cite{tancik2020stegastamp}, Stable Signature \cite{fernandez2023stable}, and Tree-Ring \cite{wen2023treering}. Each watermark family contributes equally to the aggregate results, so the reported averages are not dominated by any single method.

\subsection{Metrics}
For multi-bit methods, the reported metrics are BER, PSNR, and SSIM \cite{wang2004ssim}. BER closer to 0.5 indicates that the evaluated decoder approaches random guessing; raw BER values above 0.5 are reported but are not interpreted as stronger than the random-guessing ceiling. Stable Signature and StegaStamp are evaluated with their method-specific bit decoders. Tree-Ring is detector-based, so removal accuracy is reported at the fixed operating point as one minus TPR at the fixed low-FPR detector setting, together with PSNR and SSIM.

\subsection{Operating Curves}
The $K/U/S$ scan evaluates latent perturbation strength $K$, the controlled auxiliary keep ratio $U$ in $x_0^{(U)}$, and diffusion strength $S$. We use a coarse scan to expose the low-fidelity boundary and two finer scans around the usable region. The fine scan covers $K\in[0.85,1.10]$, $U\in\{0,0.05,0.10\}$, and $S\in[0.01,0.12]$; the finer scan covers $K\in[0.8,0.925]$, $U\in[0,0.05]$, and $S\in\{0.03,0.04,0.05,0.06,0.08\}$. To align the case study tables and the ablation panel, the fixed controlled operating point $K=0.925,U=0.02,S=0.04$ is reported for every watermark family rather than selecting a separate point for each method.

\section{Results}

Tables~\ref{tab:traditional-attack-comparison} and~\ref{tab:reconstruction-attack-comparison} test whether residual-channel suppression can act as a no-box watermark remover across heterogeneous watermark families, rather than as a decoder-specific postprocessor. The threat model follows the formal objective: the attacker has no access to the watermarking method, key, payload, or decoder. The comparison includes conventional postprocessing methods and stronger reconstruction or regeneration attacks, namely CtrlRegen, two 60-step reverse-diffusion attacks (WatermarkAttacker Diff 60 and SADRE Diff 60), and two VAE-based learned-compression attacks (BMSHJ2018 and Cheng2020) \cite{balle2018variational,cheng2020learned}. At the fixed \method{} operating point $K=0.925,U=0.02,S=0.04$, without per-method tuning, the attack obtains a mean BER of 0.400 over the six bit-decoder families, with a mean PSNR of 28.98 dB and a mean SSIM of 0.902. It also reaches a Tree-Ring removal accuracy of 0.4700 at 27.57 dB. These results put \method{} in the regime required by Definition 1 and Proposition 1, where watermark disruption and fidelity preservation must be achieved at the same time. The evidence should be read conservatively. And they show that recoverable payload evidence is weakened for the evaluated decoders while the visible semantic carrier remains within the fidelity constraint.

\begin{figure*}[!t]
\centering
\includegraphics[width=0.92\textwidth]{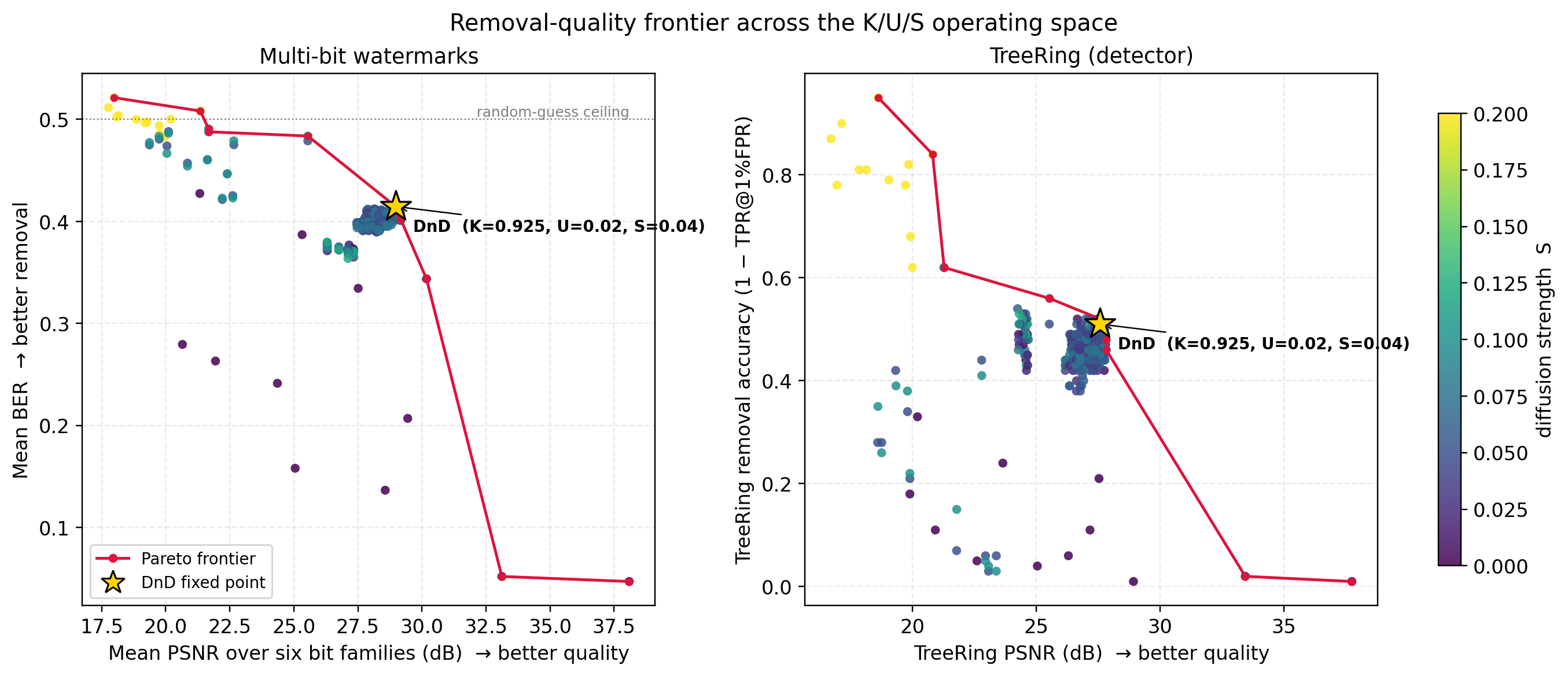}
\caption{Trade-off between BER and PSNR across attack families. The plot tests whether watermark disruption can be separated from visible image degradation. Conventional image edits can reach high removal only by moving to regimes with low PSNR, while reconstruction and regeneration baselines show method-dependent tradeoffs. The fixed controlled \method{} operating point lies in the high-fidelity region, consistent with learned residual channel suppression being distinct from visible image degradation.}
\label{fig:ber-psnr-frontier}
\end{figure*}

\begin{figure*}[!t]
\centering
\includegraphics[width=0.92\textwidth]{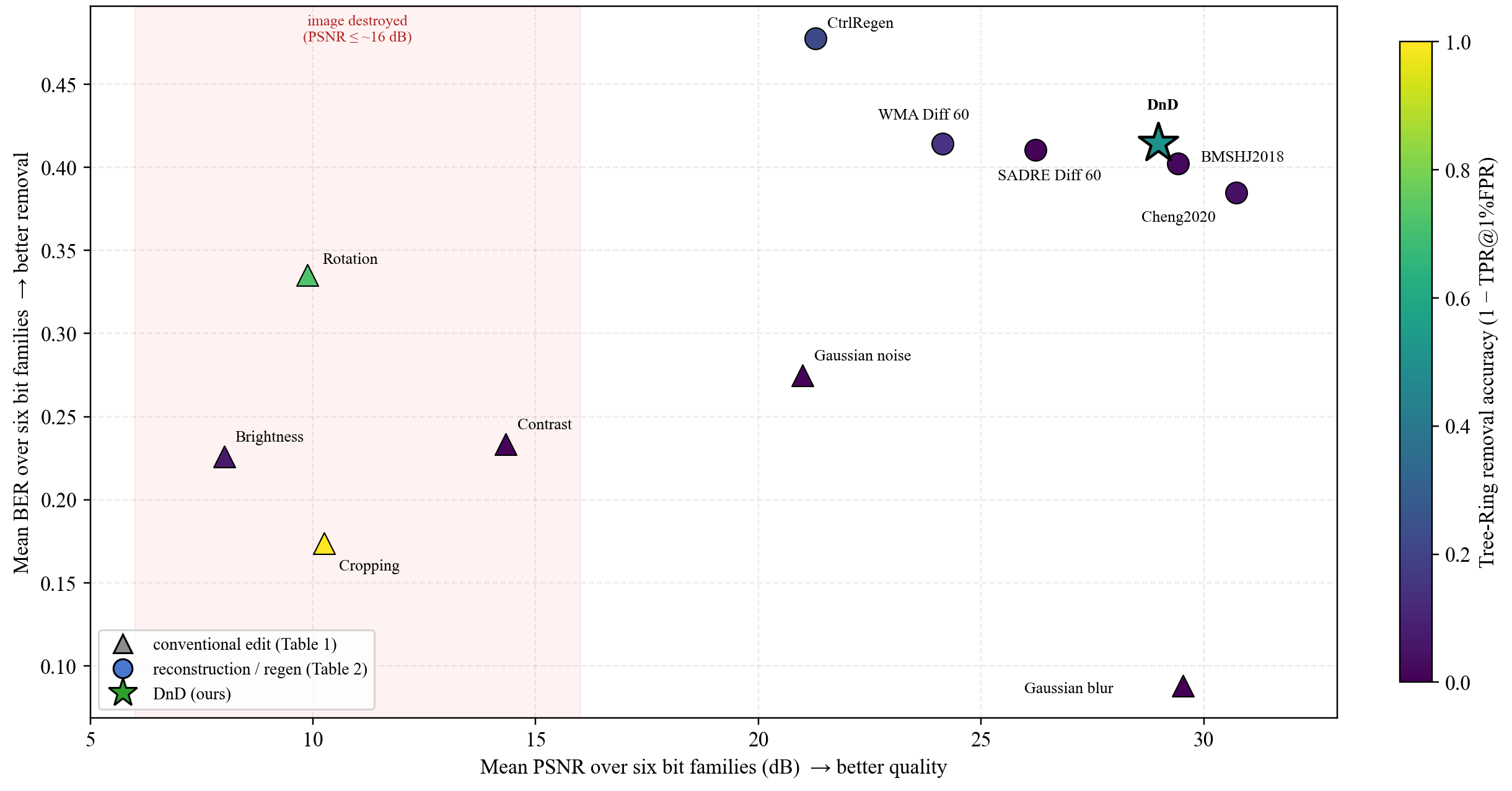}
\caption{Tables~\ref{tab:traditional-attack-comparison} and~\ref{tab:reconstruction-attack-comparison} provide an aggregate comparison of watermark removal performance and image quality across different attack endpoints. The horizontal axis reports the mean PSNR over the six bit-decoder watermark families, the vertical axis reports the corresponding mean BER, and color indicates the Tree-Ring removal accuracy under the fixed low-FPR detector setting. The shaded region denotes the low-fidelity range. The fixed \method{} operating point weakens payload recovery for the bit-decoder families while keeping the image content within the fidelity constraint, and it still retains a measurable detector-removal effect on Tree-Ring.
}
\label{fig:baseline-scatter}
\end{figure*}

\begin{table*}[!t]
  \caption{Quantitative comparison between conventional watermark removal attacks in image space and the fixed controlled operating point of \method{} ($K=0.925,U=0.02,S=0.04$). BER is reported for bit-decoder watermarks, and Removal Acc. is reported for Tree-Ring. Bold values mark the strongest raw entry within this table; BER values above 0.5 should be read under the random-guessing ceiling described in the protocol.}
  \label{tab:traditional-attack-comparison}
  \centering
  \scriptsize
  \resizebox{\linewidth}{!}{%
  \begin{tabular}{lccccccccccccccccccccc}
    \toprule
    Attack endpoint & \multicolumn{3}{c}{DwtDct} & \multicolumn{3}{c}{DwtDctSvd} & \multicolumn{3}{c}{RivaGAN} & \multicolumn{3}{c}{SSL} & \multicolumn{3}{c}{StegaStamp} & \multicolumn{3}{c}{Stable Signature} & \multicolumn{3}{c}{Tree-Ring} \\
    \cmidrule(lr){2-4} \cmidrule(lr){5-7} \cmidrule(lr){8-10} \cmidrule(lr){11-13} \cmidrule(lr){14-16} \cmidrule(lr){17-19} \cmidrule(lr){20-22}
     & BER$\uparrow$ & PSNR$\uparrow$ & SSIM$\uparrow$ & BER$\uparrow$ & PSNR$\uparrow$ & SSIM$\uparrow$ & BER$\uparrow$ & PSNR$\uparrow$ & SSIM$\uparrow$ & BER$\uparrow$ & PSNR$\uparrow$ & SSIM$\uparrow$ & BER$\uparrow$ & PSNR$\uparrow$ & SSIM$\uparrow$ & BER$\uparrow$ & PSNR$\uparrow$ & SSIM$\uparrow$ & Removal Acc.$\uparrow$ & PSNR$\uparrow$ & SSIM$\uparrow$ \\
    \midrule
    Gaussian blur & 0.3481 & 29.30 & 0.8906 & 0.0000 & \textbf{30.36} & 0.9050 & 0.0000 & \textbf{30.77} & 0.8945 & 0.0000 & 28.16 & 0.8413 & 0.0000 & 28.09 & 0.8482 & 0.1813 & \textbf{30.53} & \textbf{0.9062} & 0.0000 & \textbf{29.48} & \textbf{0.9003} \\
    Gaussian noise & \textbf{0.6038} & 20.59 & 0.4996 & 0.3025 & 20.49 & 0.4762 & 0.0516 & 20.55 & 0.4794 & \textbf{0.5055} & 20.53 & 0.5135 & 0.0018 & 20.52 & 0.5072 & 0.1842 & 23.28 & 0.4925 & 0.0100 & 22.45 & 0.4555 \\
    Brightness & 0.4591 & 7.78 & 0.2871 & 0.3787 & 8.00 & 0.2931 & 0.1391 & 8.39 & 0.2910 & 0.0023 & 8.07 & 0.2538 & 0.0000 & 7.87 & 0.2584 & 0.3762 & 7.93 & 0.2425 & 0.0700 & 7.32 & 0.2448 \\
    Contrast & 0.4587 & 14.21 & 0.6466 & 0.3787 & 14.35 & 0.6687 & 0.1309 & 14.39 & 0.6497 & 0.0048 & 14.34 & 0.5788 & 0.0000 & 14.00 & 0.5811 & 0.4279 & 14.69 & 0.5733 & 0.0100 & 13.91 & 0.5498 \\
    Rotation & 0.4097 & 9.69 & 0.2187 & 0.4094 & 9.98 & 0.2392 & 0.3081 & 9.82 & 0.2329 & 0.0333 & 9.98 & 0.2122 & 0.4227 & 9.59 & 0.2100 & \textbf{0.4281} & 10.19 & 0.2255 & 0.7200 & 8.98 & 0.2101 \\
    Cropping & 0.0356 & 10.09 & 0.2795 & 0.0000 & 10.26 & 0.2909 & 0.0013 & 10.08 & 0.2674 & 0.4070 & 10.41 & 0.1865 & \textbf{0.4860} & 9.87 & 0.1823 & 0.1119 & 10.79 & 0.1939 & \textbf{1.0000} & 9.35 & 0.2446 \\
    \midrule
    \method{} ($K=0.925,U=0.02,S=0.04$) & 0.4266 & \textbf{29.71} & \textbf{0.9301} & \textbf{0.4828} & 29.88 & \textbf{0.9340} & \textbf{0.4078} & 30.07 & \textbf{0.9235} & 0.3525 & \textbf{28.73} & \textbf{0.8861} & 0.3848 & \textbf{29.16} & \textbf{0.8989} & 0.3452 & 26.33 & 0.8405 & 0.4700 & 27.57 & 0.8826 \\
    \bottomrule
  \end{tabular}%
  }
\end{table*}

\begin{table*}[!t]
  \caption{Quantitative comparison between reconstruction or regeneration endpoints for watermark removal and the fixed controlled operating point of \method{} ($K=0.925,U=0.02,S=0.04$). WatermarkAttacker Diff 60 and SADRE Diff 60 denote 60-step reverse-diffusion reconstruction attacks, while BMSHJ2018 and Cheng2020 denote VAE-based learned-compression reconstruction endpoints \cite{balle2018variational,cheng2020learned}. BER is reported for bit-decoder watermarks, and Removal Acc. is reported for Tree-Ring. Bold values mark the strongest raw entry within this table; BER values above 0.5 should be read under the random-guessing ceiling described in the protocol.}
  \label{tab:reconstruction-attack-comparison}
  \centering
  \scriptsize
  \resizebox{\linewidth}{!}{%
  \begin{tabular}{lccccccccccccccccccccc}
    \toprule
    Attack endpoint & \multicolumn{3}{c}{DwtDct} & \multicolumn{3}{c}{DwtDctSvd} & \multicolumn{3}{c}{RivaGAN} & \multicolumn{3}{c}{SSL} & \multicolumn{3}{c}{StegaStamp} & \multicolumn{3}{c}{Stable Signature} & \multicolumn{3}{c}{Tree-Ring} \\
    \cmidrule(lr){2-4} \cmidrule(lr){5-7} \cmidrule(lr){8-10} \cmidrule(lr){11-13} \cmidrule(lr){14-16} \cmidrule(lr){17-19} \cmidrule(lr){20-22}
     & BER$\uparrow$ & PSNR$\uparrow$ & SSIM$\uparrow$ & BER$\uparrow$ & PSNR$\uparrow$ & SSIM$\uparrow$ & BER$\uparrow$ & PSNR$\uparrow$ & SSIM$\uparrow$ & BER$\uparrow$ & PSNR$\uparrow$ & SSIM$\uparrow$ & BER$\uparrow$ & PSNR$\uparrow$ & SSIM$\uparrow$ & BER$\uparrow$ & PSNR$\uparrow$ & SSIM$\uparrow$ & Removal Acc.$\uparrow$ & PSNR$\uparrow$ & SSIM$\uparrow$ \\
    \midrule
    CtrlRegen & 0.4216 & 21.16 & 0.6242 & 0.4713 & 21.58 & 0.6455 & \textbf{0.4803} & 21.66 & 0.6272 & \textbf{0.5298} & 20.83 & 0.5594 & \textbf{0.4505} & 20.91 & 0.5949 & \textbf{0.5112} & 21.57 & 0.6320 & 0.2200 & 21.68 & 0.6570 \\
    WatermarkAttacker Diff 60 & 0.4153 & 23.96 & 0.7154 & 0.4088 & 24.59 & 0.7357 & 0.4106 & 24.84 & 0.7210 & 0.4885 & 23.22 & 0.6439 & 0.2570 & 23.37 & 0.6755 & 0.5050 & 24.80 & 0.7378 & 0.1500 & 24.98 & 0.7702 \\
    BMSHJ2018 & 0.4113 & 29.30 & 0.8676 & 0.4000 & 29.96 & 0.8760 & 0.3934 & 30.15 & 0.8615 & 0.5053 & 28.31 & 0.8115 & 0.3683 & 28.41 & 0.8272 & 0.3360 & 30.34 & 0.8679 & 0.0200 & 29.91 & 0.8587 \\
    Cheng2020 & 0.4119 & \textbf{30.59} & 0.8943 & 0.3875 & \textbf{31.05} & 0.8968 & 0.4025 & \textbf{31.26} & 0.8851 & 0.4905 & \textbf{29.94} & 0.8503 & 0.3619 & \textbf{30.08} & 0.8629 & 0.2533 & \textbf{31.45} & \textbf{0.8861} & 0.0400 & \textbf{31.36} & 0.8812 \\
    SADRE Diff 60 & 0.4156 & 25.99 & 0.7663 & \textbf{0.4844} & 26.68 & 0.7844 & 0.3559 & 27.02 & 0.7647 & 0.4838 & 24.98 & 0.6906 & 0.2465 & 25.19 & 0.7237 & 0.4771 & 27.47 & 0.8087 & 0.0100 & 28.44 & 0.8559 \\
    \midrule
    \method{} ($K=0.925,U=0.02,S=0.04$) & \textbf{0.4266} & 29.71 & \textbf{0.9301} & 0.4828 & 29.88 & \textbf{0.9340} & 0.4078 & 30.07 & \textbf{0.9235} & 0.3525 & 28.73 & \textbf{0.8861} & 0.3848 & 29.16 & \textbf{0.8989} & 0.3452 & 26.33 & 0.8405 & \textbf{0.4700} & 27.57 & \textbf{0.8826} \\
    \bottomrule
  \end{tabular}%
  }
\end{table*}

\begin{figure*}[!t]
\centering
\includegraphics[width=0.98\textwidth,height=0.72\textheight,keepaspectratio]{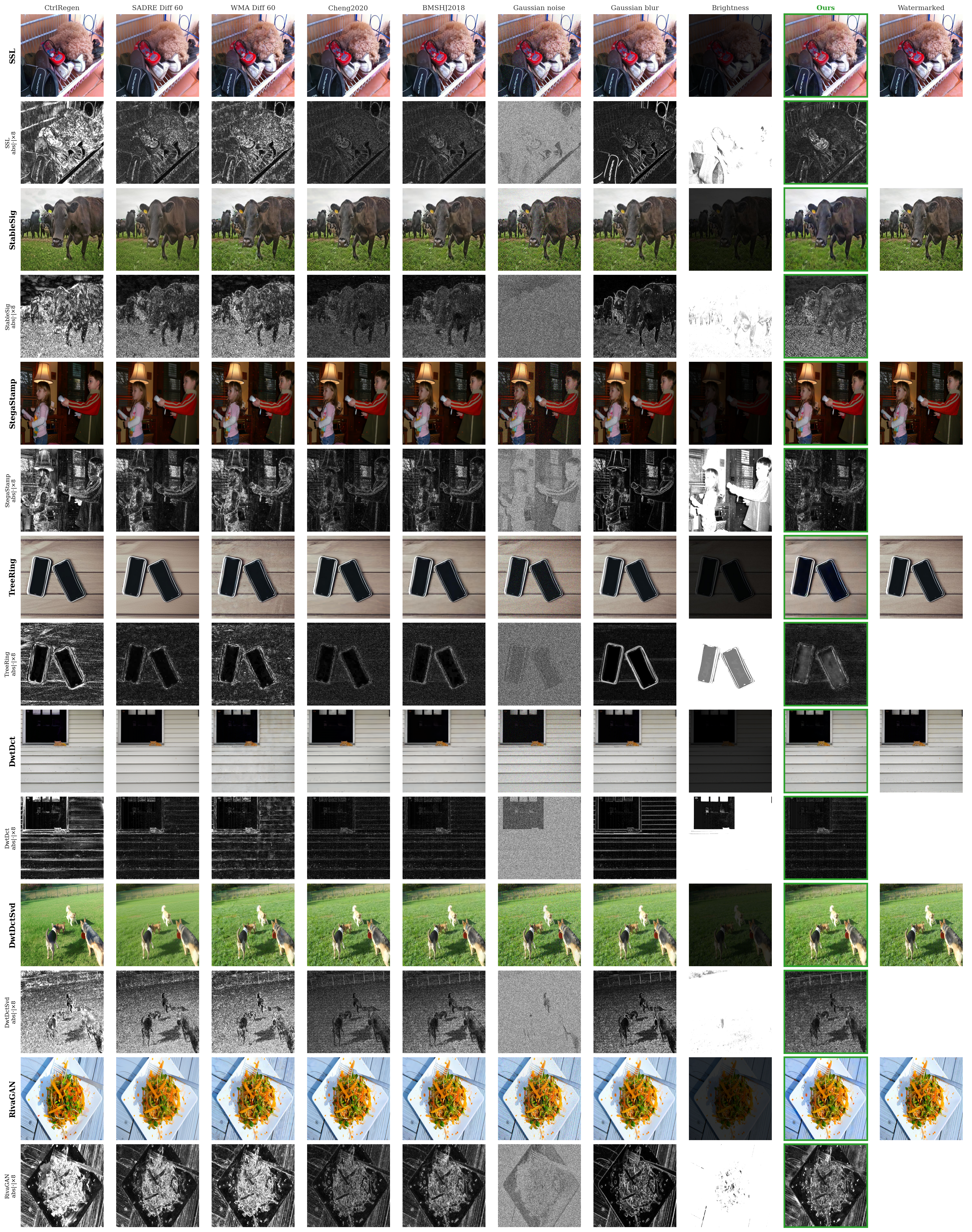}
\caption{Qualitative comparison across seven watermark families. For each watermark family, the image row shows the attacked output and the residual row shows the amplified absolute difference to the watermarked reference. The fixed \method{} endpoint is highlighted in green. Compared with severe brightness changes or high-noise attacks, \method{} preserves image content while altering residual evidence used by watermark decoders.}
\label{fig:qualitative-comparison}
\end{figure*}

Figure~\ref{fig:baseline-scatter} summarizes the trade-off captured by the two comparison tables and distinguishes effective removal from low-fidelity failure cases. Conventional edits weaken some decoders, but their high-removal cases often violate the fidelity constraint: brightness adjustment, rotation, cropping, and contrast changes reduce PSNR to about 7--15 dB, while Gaussian noise degrades structural fidelity even when it drives some raw BER values above the random-guessing level. Gaussian blur has the opposite limitation: it preserves fidelity but gives limited removal across methods. Reconstruction and regeneration endpoints show a similar trade-off. The 60-step reverse-diffusion attacks, WatermarkAttacker Diff 60 and SADRE Diff 60, weaken bit decoders, but at substantially lower fidelity. The VAE-based learned-compression endpoints, BMSHJ2018 and Cheng2020, better preserve image content but remove little detector evidence from Tree-Ring. The fixed \method{} point is not optimal for every metric, yet it occupies a more useful part of the trade-off. It weakens recoverable payload evidence across transform-domain, learned, and generative watermark families while keeping image content within the fidelity constraint. This behavior matches the mechanism in Section~III: the encoder routes visible structure into $g$, the auxiliary branch $u$ is suppressed, and the decoder reconstructs from the routed carrier rather than overwriting the image. The security implication is that robustness to conventional edits is not enough; watermark evaluation should also include learned representation-level removers that suppress residual evidence without visibly damaging the image. Figure~\ref{fig:qualitative-comparison} gives the corresponding visual comparison.

\subsection{Operating and Ablation Evidence}
Table~\ref{tab:scan-ablation-evidence} links the aggregate removal results to the representation-routing design through parameter scans and component ablations. The BER, PSNR, and SSIM columns are averaged over the six bit-decoder watermark families, while Tree-Ring is reported separately because it is evaluated with a detector-based metric. The three blocks address related questions about the mechanism: where the useful $K/U/S$ regime lies, whether adding $u$ back restores image content or payload evidence, and which training or inference components are needed to suppress residual evidence without violating the fidelity constraint. This organization helps separate design-driven removal from incidental image damage.

\begin{table*}[!t]
\caption{Aggregated scan and ablation evidence. BER, PSNR, and SSIM average SSL, DwtDct, DwtDctSvd, RivaGAN, Stable Signature, and StegaStamp; Tree-Ring removal accuracy is reported separately.}
\label{tab:scan-ablation-evidence}
\centering
\small
\setlength{\tabcolsep}{4pt}
\resizebox{\linewidth}{!}{%
\begin{tabular}{llcccc}
\toprule
Setting & $K/U/S$ or Slice & BER$\uparrow$ & PSNR$\uparrow$ & SSIM$\uparrow$ & Tree-Ring Removal Acc.$\uparrow$ \\
\midrule
\multicolumn{6}{l}{A. K/U/S scan evidence} \\
Coarse scan: maximum removal boundary & $1.5/0.5/0.2$ & 0.521 & 17.99 & 0.464 & 0.90 \\
Coarse scan: best high-fidelity point, PSNR $\geq 28$ & $0.5/0/0$ & 0.344 & 30.18 & 0.932 & 0.21 \\
Fine scan: best bit removal & $0.85/0/0.04$ & 0.411 & 28.43 & 0.900 & 0.51 \\
Fine scan: best bit removal, PSNR $\geq 29$ & $1.0/0/0.10$ & 0.408 & 29.17 & 0.905 & 0.46 \\
Finer scan: best bit removal & $0.825/0/0.03$ & 0.412 & 28.14 & 0.899 & 0.45 \\
Finer scan: best bit removal, PSNR $\geq 29$ & $0.925/0/0.05$ & 0.409 & 29.04 & 0.904 & 0.46 \\
\midrule
\multicolumn{6}{l}{B. Does $U$ recover images after diffusion?} \\
Coarse scan: no diffusion, residual removed & mean over $K$, $U=0,S=0$ & 0.373 & 26.09 & 0.839 & 0.22 \\
Coarse scan: no diffusion, residual restored & mean over $K$, $U=1,S=0$ & 0.098 & 31.21 & 0.866 & 0.02 \\
Coarse scan: diffusion active, no residual & mean over $K,S>0$, $U=0$ & 0.488 & 21.45 & 0.666 & 0.61 \\
Coarse scan: diffusion active, residual restored & mean over $K,S>0$, $U=1$ & 0.456 & 21.37 & 0.672 & 0.28 \\
Finer scan: low $S$ diffusion/refiner, $U=0$ & mean over $K,S$, $U=0$ & 0.408 & 28.50 & 0.901 & 0.46 \\
Finer scan: low $S$ diffusion/refiner, larger $U$ & mean over $K,S$, $U=0.05$ & 0.394 & 27.95 & 0.896 & 0.44 \\
Fine scan: low $S$ diffusion/refiner, $U=0$ & mean over $K,S$, $U=0$ & 0.406 & 28.84 & 0.903 & 0.48 \\
Fine scan: low $S$ diffusion/refiner, larger $U$ & mean over $K,S$, $U=0.10$ & 0.372 & 27.00 & 0.895 & 0.49 \\
\midrule
\multicolumn{6}{l}{C. Component ablation at $K=0.925$, $U=0.02$, $S=0.04$} \\
Full model & $0.925/0.02/0.04$ & 0.400 & 28.98 & 0.902 & 0.47 \\
w/o Frequency Aware Latent Attack & $0.925/0.02/0.04$ & 0.495 & 24.40 & 0.788 & 0.61 \\
w/o Clean Drop Supervision & $0.925/0.02/0.04$ & 0.449 & 22.80 & 0.749 & 0.23 \\
w/o Stage 3 Diffusion Exposure & $0.925/0.02/0.04$ & 0.409 & 28.57 & 0.881 & 0.43 \\
Unfrozen Stage 2 Base & $0.925/0.02/0.04$ & 0.371 & 31.10 & 0.945 & 0.41 \\
w/o Test Time Refiner & $0.925/0.02/0.04$ & 0.363 & 25.80 & 0.883 & 0.28 \\
\bottomrule
\end{tabular}%
}
\end{table*}

\subsection{Attack Operating Regime}
The $K/U/S$ scan separates the practical operating region from the aggressive end of the search. In our design, $K$ applies frequency-aware latent perturbation to the structure carrier, $U$ controls how much auxiliary residual evidence is reintroduced, and $S$ controls the post-drop diffusion stress. The coarse scan identifies the low-fidelity boundary: the maximum-removal setting reaches a mean BER of 0.521 and a Tree-Ring removal accuracy of 0.90, but only with a PSNR of 17.99 dB and an SSIM of 0.464. This endpoint is useful for locating the failure boundary of the watermarking methods, but it is not a viable removal point under a fidelity constraint. The fine and finer scans instead locate the practical regime. With low $U$ and a small positive $S$, BER stays close to 0.41 while PSNR remains around 28--29 dB, and the PSNR-constrained rows show a similar removal level. This pattern is consistent with Corollary 1: aggressive image-space or diffusion perturbations can raise removal scores, but useful removal comes from controlling the latent residual. The fixed shared operating point is therefore selected from the practical regime rather than from the maximum-removal boundary, so the comparison reflects cross-family attack capability without per-method oracle tuning.

\subsection{Auxiliary Residual Reintroduction Under Diffusion}
The controlled $U$ scan clarifies the role of the auxiliary branch during reconstruction and after diffusion stress. Without diffusion, increasing $U$ restores residual information: moving from $U=0,S=0$ to $U=1,S=0$ raises PSNR from 26.09 dB to 31.21 dB, but lowers the mean BER from 0.373 to 0.098 and almost eliminates the detector-removal effect on Tree-Ring. This matches the information gate in Proposition 1. The full latent pair reconstructs the watermarked observation more faithfully, whereas the latent pair with $u$ dropped removes evidence used by the evaluated decoder or detector. Once diffusion is active, reintroducing the same residual no longer improves fidelity. In the coarse $S>0$ scan, PSNR changes only from 21.45 dB to 21.37 dB as $U$ increases from 0 to 1, while Tree-Ring removal accuracy falls from 0.61 to 0.28. The low-$S$ scans show the same trend: larger $U$ reduces removal without improving PSNR. These results suggest that $u$ contains useful image residuals before regeneration, but after diffusion stress it can also reintroduce detector- or decoder-relevant evidence. A practical attack should therefore suppress $u$ and repair the carrier after $u$ is dropped, rather than treating residual reintroduction as a general image-recovery mechanism.

\begin{figure*}[!t]
\centering
\includegraphics[width=0.90\textwidth]{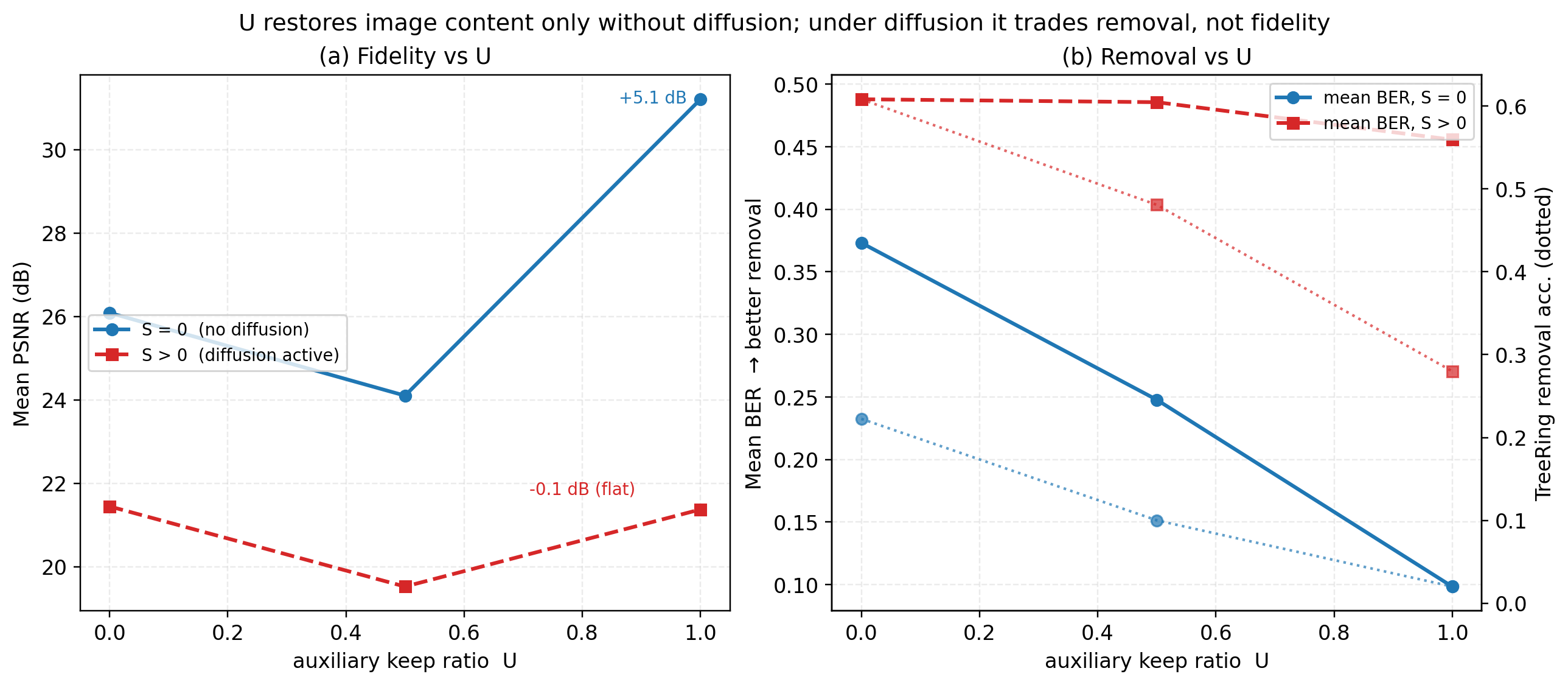}
\caption{Effect of reintroducing the auxiliary residual branch under different diffusion regimes. Without diffusion, larger $U$ behaves like residual restoration and improves PSNR while reducing removal. Once diffusion is active, larger $U$ no longer recovers fidelity and mainly weakens the attack, which supports the interpretation that strong diffusion is a poor core mechanism for useful watermark removal.}
\label{fig:u-diffusion}
\end{figure*}

\subsection{Component Contributions}
The ablations further explain why residual suppression under a fidelity constraint requires a controlled separation between $g$ and $u$, followed by inference-time refinement.Replacing the frequency-aware latent attack with latent noise increases mean BER to 0.495 and Tree-Ring removal accuracy to 0.61, but the PSNR drop to 24.40 dB violates the fidelity constraint. This indicates that the frequency-aware attack is not designed to maximize perturbation magnitude; it regularizes the structure carrier across spectral bands so that $g$ remains useful for reconstruction after $u$ is suppressed. Removing clean-drop supervision gives a similar failure mode: BER rises to 0.449, but PSNR collapses to 22.80 dB and Tree-Ring removal accuracy falls to 0.23. The cause is that the structure branch becomes an unconstrained reconstruction path, which breaks the intended separation between visible content in $g$ and residual evidence in $u$. Removing Stage 3 diffusion exposure slightly increases mean BER but lowers PSNR and detector removal, indicating that diffusion exposure trains stability against adaptive reconstruction rather than serving as the main remover. Unfreezing the Stage 2 base shifts the model toward reconstruction, improving PSNR to 31.10 dB but reducing mean BER to 0.371 and Tree-Ring removal accuracy to 0.41; this is consistent with the base network relearning residual recovery instead of enforcing the drop path. Disabling the test-time refiner weakens both removal and fidelity, which shows that the refiner improves the post-drop structure using local restoration and attention context without simply restoring the payload-bearing residual branch. Together, these ablations show that useful removal depends on placing semantic content in $g$, suppressing residual evidence in $u$, stress-testing the drop path through frequency and diffusion controls, and refining the output only after the residual channel has been reduced.

\begin{figure*}[!t]
\centering
\includegraphics[width=0.92\textwidth]{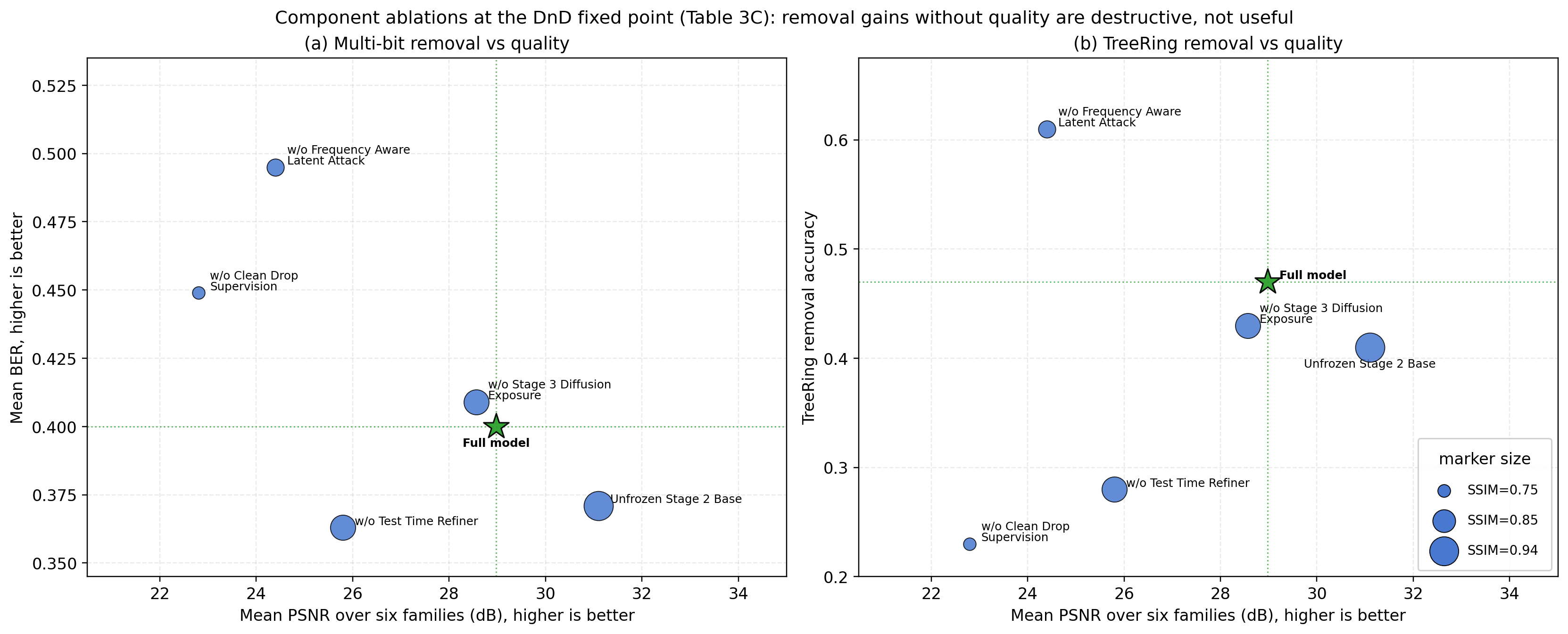}
\caption{Ablation tradeoff at $K=0.925,U=0.02,S=0.04$. The scatter view separates useful removal under a fidelity constraint from low-fidelity variants with high raw removal scores. Removing the frequency-aware latent attack or clean-drop supervision increases removal metrics at substantial fidelity cost, whereas unfreezing the Stage 2 base shifts the model toward reconstruction and disabling the test-time refiner shows the role of post-drop repair.}
\label{fig:ablation-scatter}
\end{figure*}

\section{Discussion}

\subsection{Implications for Watermark Design}
\method{} suggests that a watermark can be vulnerable when its payload can be routed into a low-energy, discardable residual channel without disturbing the semantic carrier. This complements recent robustness benchmarks and removal studies showing that imperceptible persistence under standard image edits does not imply resistance to adaptive reconstruction \cite{an2024waves,zhao2024provably,hu2024stableunstable,alam2025sadre}. Defenses should not rely only on residual persistence or on robustness to standard postprocessing suites. More robust designs may need to bind payload evidence to semantic structure, randomize or key the residual pattern across images, distribute it across content factors that cannot be discarded, and combine watermarking with cryptographic provenance rather than treating any single invisible signal as sufficient.

\subsection{The Limited Role of Strong Diffusion in Removal}
Diffusion regeneration can weaken hidden perturbations, but Table~\ref{tab:scan-ablation-evidence} shows that it is not a reliable recovery mechanism for fidelity-constrained watermark removal. Once diffusion is active, reintroducing the auxiliary residual through larger $U$ does not restore PSNR; it mainly reduces removal strength. This matches the behavior of watermark attacks based on regeneration: stronger reconstruction can raise BER by overwriting image evidence, but the gain is often paid for by visible distortion rather than targeted watermark suppression \cite{zhao2024provably,liu2025ctrlregen,alam2025sadre}. In \method{}, diffusion is best understood as an adaptive stressor for training and as a low-strength refinement component at inference. The useful removal behavior comes from latent drop: preserve $g$, weaken $u$, and avoid using image destruction as the main source of removal.

\subsection{Diffusion as Semantic Regeneration Rather Than Reconstruction Attack}

Diffusion models are effective semantic regeneration tools, but they are not ideal core mechanisms for fidelity-constrained watermark removal. Their strength lies in redrawing an image on the natural-image manifold while preserving coarse semantics, object layout, and structural content. This behavior is useful for image editing, purification, and content-preserving regeneration, where a plausible regenerated image is often acceptable. A watermark removal attack has a stricter objective: it must reduce decoder- or detector-relevant watermark evidence while keeping the visible carrier close to the original image under a fidelity constraint.

This distinction explains why strong diffusion should not be treated as the main reconstruction attack in this setting. Strong diffusion can increase BER or reduce detector response by overwriting low-level residual evidence, but the same process may also modify non-watermark details and move the output into a low-fidelity regime. Such a result is better understood as aggressive regeneration rather than useful removal. Low-strength diffusion has a different role. It can serve as a stress and repair component after latent drop, testing whether residual evidence reappears under regeneration and helping refine the post-drop image. The useful attack mechanism in \method{} therefore comes from representation-level residual suppression: preserve the structure latent $g$, suppress the auxiliary residual latent $u$, and avoid relying on image rewriting as the source of watermark removal.

\section{Conclusion}

We studied whether invisible watermark robustness remains meaningful when the attacker is a learned remover at the representation level rather than a conventional image editor. The results show that robustness to compression, resizing, blur, or color changes is not sufficient: a payload that survives benign postprocessing may still be routed into a discardable residual channel and removed while preserving the semantic image carrier.

\method{} frames removal as a reconstructive grayscale residual routing problem. The experiments identify a practical attack regime in which \method{} transfers across seven watermark families under one shared setting, while operating scans show that stronger noise or diffusion mainly improves removal by damaging the image. The ablation study further confirms that useful removal requires quality-aware latent dropping and residual repair, not simply more aggressive perturbation.

Future work should examine learned removers at the representation level under broader datasets, unseen watermark keys, and stronger adaptive watermark designs. More broadly, this study raises a practical question for provenance systems: how should robustness be evaluated when the attacker can learn to separate semantic content from removable residual evidence?

\section*{Acknowledgments}
This work was supported in part by Taishan Scholar under Grant tsqnz20250747; in part by the National Natural Science Foundation under Grant 62502250, Grant 62406051, Grant 62302249, Grant 62541206, and Grant 62272255; and in part by the Young Talent of Lifting Engineering for Science and Technology in Shandong under Grant SDAST2025QTB030.


\end{document}